%% file: root.tex
\documentclass[letterpaper, 10 pt, journal, twoside]{IEEEtran}  

\IEEEoverridecommandlockouts                              

\usepackage{graphics} 
\usepackage{amsmath} 
\usepackage{amssymb}  
\usepackage{capt-of}
\usepackage{lipsum}
\usepackage{cuted}
\usepackage{stfloats}
\usepackage{makecell}
\usepackage{tabu, booktabs}
\usepackage{colortbl}
\usepackage{xcolor}

\usepackage{algorithm, algorithmic}
\usepackage{graphicx}
\usepackage{tikz}
\usepackage{calc}

\usepackage{multirow}
\usepackage{arydshln}

\usepackage{caption}
\DeclareCaptionLabelSeparator{dottilde}{.~~}
\title{\LARGE \bf VG-Mapping: Variation-aware Density Control for Online 3D Gaussian Mapping in Semi-static Scenes
}
\author{Yicheng He$^{1}$, Jingwen Yu$^{1,2}$, Guangcheng Chen$^{1}$ and Hong Zhang$^{1*}$~\IEEEmembership{Life~Fellow,~IEEE}
\thanks{$^{*}$ corresponding author: Hong Zhang ({\tt \small hzhang@sustech.edu.cn})}
\thanks{$^{1}$ Southern University of Science and Technology, Shenzhen, China}
\thanks{$^{2}$ CKS Robotics Institute, Hong Kong University of Science and Technology, Hong Kong SAR, China}
}

\begin{document}
\maketitle
\thispagestyle{empty}
\pagestyle{empty}
\begin{strip}
\vspace{-2.5cm}
    \begin{center}
        \includegraphics[width=\textwidth]{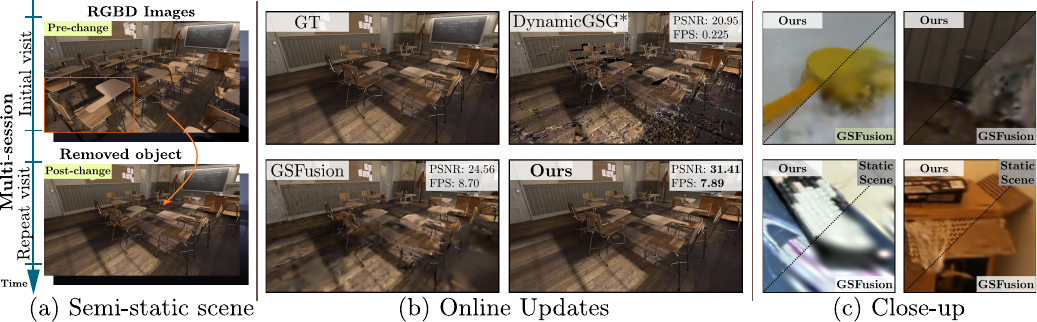}
        \captionof{figure}{We propose VG-Mapping, an RGB-D online 3DGS mapping system tailored to  semi-static scenes. (a) When a robot revisits the same place, dynamic changes across visits may cause inconsistencies between the prior map and the current observation, which we define as semi-static regions. To address this issue, VG-Mapping introduces a variation-aware mapping mechanism that (b) efficiently and accurately updates the changed areas. 
        (c) Although our framework is primarily designed for semi-static environments, it also brings improvements in static setting vs. GSFusion, a state-of-the-art competing method.}
        \label{fig:highlight}
    \end{center}
\vspace{-0.5cm}
\end{strip}

\input{chapter/abstract}
\input{chapter/introduction}
\input{chapter/related_works}
\input{chapter/method}
\input{chapter/experiments}
\input{chapter/conclusion}

\bibliographystyle{IEEEtran}
\bibliography{IEEEabrv, ref}
\end{document}

%% file: chapter/abstract.tex
\begin{abstract}
Maintaining an up-to-date map that accurately reflects recent changes in the environment is crucial, especially for robots that repeatedly traverse the same space. Failing to promptly update the changed regions can degrade map quality, resulting in poor localization, inefficient operations, and even lost robots. 3D Gaussian Splatting (3DGS) has recently seen widespread adoption in online map reconstruction due to its dense, differentiable, and photorealistic properties, yet accurately and efficiently updating the regions of change remains a challenge. In this paper, we propose VG-Mapping, a novel online 3DGS-based mapping system tailored for such semi-static scenes. 
Our approach introduces a variation-aware density control strategy that decouples Gaussian density regulation from optimization. Specifically, we identify regions with variation to guide initialization and pruning, which avoids the use of stale information in defining the starting point for the subsequent optimization.
Furthermore, to address the absence of public benchmarks for this task, we construct a RGB-D dataset comprising both synthetic and real-world semi-static environments. Experimental results demonstrate that our method substantially improves the rendering quality and map update efficiency in semi-static scenes. 
The code and dataset are available at https://github.com/heyicheng-never/VG-Mapping.
\end{abstract}


%% file: chapter/introduction.tex
\section{Introduction}
\IEEEPARstart{S}{imultaneous} Localization and Mapping (SLAM) systems are widely applied in robotics, AR/VR. Traditional SLAM algorithms~\cite{campos2021orb} primarily emphasize localization, with mapping serving only as a means to support it. This results in sparse maps that are inadequate to support downstream applications beyond localization that require a dense map representation. Radiance fields, a recently proposed map representation, provide a differentiable formulation for both mapping and rendering. It uses end-to-end optimization to learn dense scene representations without relying on scarce 3D ground-truth data. Among the proposed approaches, Neural Radiance Fields (NeRF)~\cite{mildenhall2021nerf} and 3D Gaussian Splatting (3DGS)~\cite{kerbl20233d} are the most representative. Compared with NeRF, 3DGS offers faster optimization, higher-quality rendering, and greater flexibility for editing, making it a particularly attractive solution to dense mapping. In fact, 3DGS has recently been integrated into SLAM frameworks to overcome the limitations of sparse maps~\cite{matsuki2024gaussian} and serves as a map representation for downstream robotic applications.

In real-world deployments, robots frequently revisit the same place time and again. During a repeated visit, scenes may have undergone dynamic changes due to human activities, like the rearrangement of furniture, resulting in what is known as a semi-static scene, as illustrated in Fig.~\ref{fig:highlight} (a). From the robot’s perspective, a semi-static scene can be regarded as a composition of multiple snapshots captured at different instants of time. The prior map of the scene is constructed from all previously observed snapshots whereas discrepancies can exist between the prior map and robot's current sensor observations in regions that have changed. Note that semi-static scenes differ fundamentally from dynamic scenes in two ways: 1) they involve unobserved transitions, where only the result of object displacement is captured rather than the process itself; and 2) they exhibit category-agnostic changes, as movable objects can belong to diverse categories beyond animals and vehicles.
Owing to these characteristics, existing mapping strategies designed for dynamic environments~\cite{zheng2025wildgs} are not directly applicable to semi-static scenarios. These challenges highlight the need for mapping systems that can accurately update maps in real time under semi-static conditions, so as to support downstream robotic tasks.

We now return to vanilla 3DGS. In this framework, scene reconstruction is achieved by backpropagating photometric loss to optimize the attributes of Gaussian primitives, while adaptive density control (ADC) enables densification and pruning of the primitives. 
ADC refines the number of Gaussian primitives based on optimization states such as mean gradients and opacity values. While this strategy is effective for static scenes or offline global optimization, it is less suitable for online mapping in semi-static scenes. In particular, due to its slow response to scene changes, ADC fails to promptly adjust the Gaussian representation in changing regions, leading to missing primitives for newly added content and erroneous ones for removed structures. As a result, the optimization at worst converges to suboptimal solutions and requires at best substantially more iterations to compensate for inconsistencies in the changed regions.
Addressing this limitation, recent works have explored enhancing the local update capability of 3DGS. These works typically leverage vision foundation models~\cite{oquab2023dinov2, kirillov2023segment} to compare the rendered and observed images from the same viewpoint, generating 2D change masks.
These masks are subsequently leveraged to accelerate the optimization process through voting strategies~\cite{ackermann2025cl}, mask-based loss functions~\cite{zeng2025gaussianupdate}, or object-level operations on Gaussian primitives~\cite{ge2025dynamicgsg}.
Despite their progress, several challenges remain. First, since these methods still rely on optimization-based ADC, they introduce outdated Gaussian primitives from changed regions into the optimization process, which significantly limit the map accuracy at the changed areas. Second, in the absence of depth information, which is commonly available on robotic platforms, supervising 3D change regions typically relies on batch processing, thereby limiting the feasibility of real-time online updates.

Building on the above discussions, we propose VG-Mapping, an RGB-D online 3DGS mapping system tailored for semi-static environments.
In such settings, a key challenge lies in maintaining a consistent and well-initialized Gaussian representation to account for scene changes. We argue that effective semi-static online mapping critically depends on decoupling density initialization from the optimization process, allowing newly added or removed structures to be handled with appropriate initial conditions.
To enable this decoupling, VG-Mapping introduces a variation-aware density control (VDC) mechanism that introduces the critical step of variation detection to identify inconsistencies between incoming observations and the existing map, thereby guiding targeted density initialization. While 3DGS offers photorealistic appearance modeling, its lack of geometric precision hinders effective use of depth cues for detecting geometric variations. Moreover, performing 3D geometric differencing for variation detection within such unstructured, point-like representation incurs significant computational overhead~\cite{hu2024ms}.
To address these limitations, we adopt a hybrid map representation that augments the 3DGS map with a TSDF-based voxel map, providing fine-grained geometric support for reliable variation detection and efficient density initialization.
Finally, to address the lack of public RGB-D datasets for online 3DGS mapping in semi-static scenes, we construct a new benchmark comprising six simulated sequences and three real-world sequences. Experimental results demonstrate that VG-Mapping effectively exploits depth information to perform high-quality and efficient update of the 3DGS map in the changed regions of semi-static environments, as illustrated in Fig.~\ref{fig:highlight} (c).

%% file: chapter/related_works.tex
\section{Related Work}
\subsection{Map Representation}
SLAM as an indispensable perception module in robotic systems, has been extensively studied over the years. Common map representations include sparse feature-based methods such as ORB-SLAM~\cite{campos2021orb}, point cloud representations~\cite{hu2024ms}, and dense volumetric methods such as occupancy grids~\cite{hornung2013octomap} and truncated signed distance functions (TSDFs)~\cite{newcombe2011kinectfusion}. Among these, TSDFs have drawn significant attention due to their regular grid structure, rich surface information, robustness to sensor noise, ease of incremental updates, and suitability for parallel computation. However, existing representations are often either overly sparse or lack the photorealistic fidelity required for tasks such as object search and path planning. Recently, radiance fields have emerged as a differentiable and dense alternative for scene mapping, with NeRF~\cite{mildenhall2021nerf} and 3DGS~\cite{kerbl20233d} being the most representative. 3DGS models a scene with 3D Gaussian primitives and employs differentiable rasterization along with photometric loss to optimize these primitives via backpropagation, enabling high-fidelity novel view synthesis. Owing to its differentiability, dense representation, photorealistic rendering, and editability, 3DGS has rapidly gained traction as a map representation. In this work, we leverage the complementary strengths of TSDFs and 3DGS to construct a hybrid representation that simultaneously captures accurate scene geometry and photorealistic appearance, and further use the TSDF-provided geometry to guide efficient and accurate density control in 3DGS.

\subsection{Gaussian-based RGB-D Online Dense Mapping}
3DGS has been increasingly integrated into online dense mapping systems owing to its advantages.  
Active3D~\cite{li2025active3d} introduce a hybrid implicit–explicit representation that fuses neural fields with Gaussian primitives to jointly capture global structural priors and locally observed details.
GSFusion~\cite{wei2024gsfusion} incorporates a TSDF-based representation to capture surface geometry and applies a quadtree image segmentation strategy to regulate the number of initialized Gaussians.
Since these early 3DGS based SLAM algorithms assume a static scene, subsequent works have targeted dynamic scenes~\cite{zheng2025wildgs,xu2024dg,li20254d,guo20254d3r}, aiming to mitigate the influence of moving objects on static part reconstruction.
However, these methods remain ill-suited for semi-static scenes, which are characterized by unobserved transitions and category-agnostic changes that are typically not caused by people or vehicles. In this work, we focus on the commonly encountered semi-static environments in robotics and propose a Gaussian-based RGB-D online dense mapping system tailored to such scenarios.

\subsection{Local Update of 3D Gaussians}
For semi-static scenes, the most effective strategy for map updating is to perform local updates only in the regions that have changed~\cite{ali2026probper,jiao2025litevloc}. CL-Splats~\cite{ackermann2025cl} and GaussianUpdate~\cite{zeng2025gaussianupdate} exploit vision foundation models to generate 2D change masks, thereby restricting joint optimization to the changed areas. However, due to their limited use of depth information, these methods typically depend on multi-view observations or optimization cues, such as gradients or loss values, to identify the regions that require updating, which makes them inappropriate for online applications. Also using semantic-level information, DynamicGSG~\cite{ge2025dynamicgsg} first segments a scene into objects and maintains a separate set of Gaussian primitives for each object, so that subsequent updates can be carried out at the object level. Because DynamicGSG~\cite{ge2025dynamicgsg} places strong demands on object segmentation and cross-frame matching, it is prone to erroneous deletion or incomplete updates; moreover, maintaining the object set is more time-consuming than pixel-level local updates for the purpose of map maintenance. Beyond these limitations, the aforementioned methods all rely on optimization-based ADC and consequently inherit its weaknesses mentioned previously.

%% file: chapter/method.tex
\section{Methodology}
\begin{figure*}[t]
    \centering
    \includegraphics[width=0.95\textwidth]{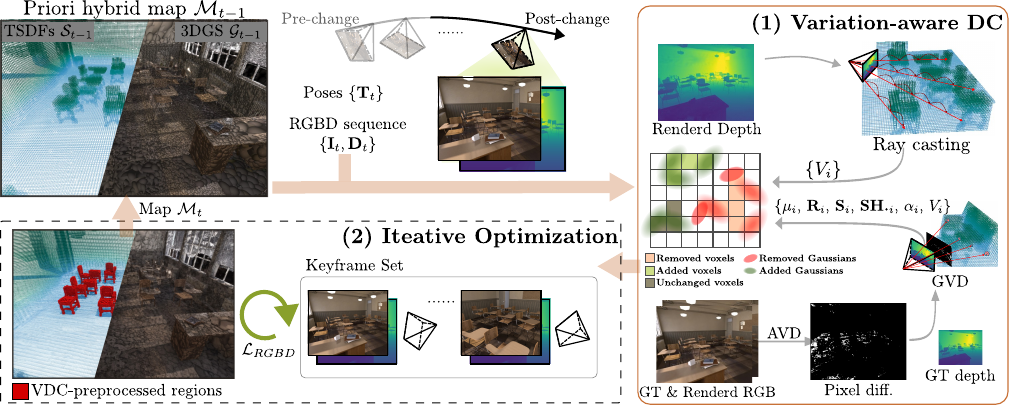}
    \caption{\textbf{VG-Mapping System Pipeline}. In terms of map representation, VG-Mapping combines 3DGS with TSDFs. We further design a variation-aware density control (VDC) module, consisting of initialization and pruning, to replace the optimization-driven adaptive density control used in vanilla 3DGS. The initialization stage leverages appearance-based (AVD) and geometry-based (GVD) variation detection to efficiently insert new Gaussian primitives.}
    \label{fig:pipeline}
    \vspace*{-0.12in}
\end{figure*}

As is typical of a modern SLAM algorithm, our system is composed of two parallel threads: tracking and mapping. The tracking thread employs VINS-Mono~\cite{qin2018vins} for real-time camera pose estimation. Since VINS-Mono~\cite{qin2018vins} operates in a frame-to-frame manner, its localization performance is unaffected by the changes in a semi-static environment. As a result, we only need to focus on the mapping thread.
The pipeline of our proposed VG-Mapping is shown in Fig.~\ref{fig:pipeline}. Given 3DGS map $\mathcal{G}_{t-1}$ at time step $t{-}1$, along with the RGB-D observation $\{\mathbf{I}_t, \mathbf{D}_t\}$ and the corresponding camera pose $\mathbf{T}_t$ at time $t$ provided by the tracking thread, our goal is to efficiently update the regions of the map that have changed—specifically, those where the current observation is inconsistent with $\mathcal{G}_{t-1}$.
This ensures that the map maintained by the robot remains up-to-date, so that changes in the environment do not adversely affect downstream tasks such as global localization, obstacle avoidance, and object search.
We will introduce the hybrid map in Sec.~\ref{sec:rep_secene} and the design of variation-aware density control (VDC) in Sec.~\ref{sec:gadc}.

\subsection{Hybrid Scene Representation}\label{sec:rep_secene}
To enhance the ability of 3DGS maps to capture fine-grained geometric information, we construct a hybrid representation by combining a TSDF-based voxel map $\mathcal{S}$ with a 3DGS map $\mathcal{G}$. Inspired by KinectFusion~\cite{newcombe2011kinectfusion}, we integrate surface measurements derived from per-frame depth images to build the global TSDF voxel map. For the 3DGS component, we optimize the Gaussian primitives via backpropagation using a loss function constructed from both RGB and depth images.

\subsubsection{TSDF-based Voxel Map}
For a given voxel size $s$, we construct the surface voxel map using an octree structure~\cite{vespa2018efficient}, which accelerates voxel access. To reduce unnecessary memory usage and computational overhead, we allocate voxels only in the vicinity of surfaces. For each voxel located at position $\mathbf{p} \in \mathrm{R}^3$, two components are stored: the truncated signed distance value $F(\mathbf{p})$ and weight $W(\mathbf{p})$,
\begin{equation}
\mathbf{S}(\mathbf{p})~~\mapsto~~[F(\mathbf{p}), W(\mathbf{p})].
\end{equation}
Given depth image $\mathbf{D}_t$ and camera pose $\mathbf{T}_t \mapsto [\mathbf{R}_t, \mathbf{t}_t]$ at time $t$, the value of voxel $\mathbf{S}_t(\mathbf{p}) \mapsto [F_t(\mathbf{p}), W_t(\mathbf{p})]$ for each $\mathbf{p}$ with in the current view frustum is computed as:
\begin{gather}
F_t(\mathbf{p})~~=~~\phi(\mathbf{D}[\mathbf{u}] - \frac{||\mathbf{t}_t - \mathbf{p}||_2}{||\mathbf{K}^{-1}(\mathbf{u}^{\mathrm{T}}|1)^\mathrm{T}||_2}),\\
\mathbf{u}=\pi(\mathbf{K}\mathbf{T}_t^{-1}(\mathbf{p}^\mathrm{T}|1)^\mathrm{T}),~~
\phi(x)=\max(-1, \min(1, \frac{x}{\mu})), \\
W_t(\mathbf{p})~~=~~\left\{
\begin{aligned}
&1, &|F_t(\mathbf{p}) - F(\mathbf{p})| < \epsilon_F,\\
&-5, &otherwise.
\end{aligned}
\right.
\end{gather}
where $\mathbf{K}$ is camera intrinsics, $\mu$ is truncated value and $\pi(\cdot)$ is perspective projection including dehomogenisation. To enhance the TSDF's responsiveness to changes in semi-static scenes, the weight is set to a negative value with a greater absolute magnitude when the difference between the TSDF value $F_t(\mathbf{p})$ computed from the current observation and the TSDF value $F(\mathbf{p})$ in the global map exceeds a threshold $\epsilon_F$.

The global $\mathbf{S}(\mathbf{p})$ is updated through $\mathbf{S}_t(\mathbf{p})$ as follows:
\begin{align}
F(\mathbf{p})~~&=~~\frac{|W_t(\mathbf{p})|F_t(\mathbf{p}) + W(\mathbf{p})F(\mathbf{p})}{|W_t(\mathbf{p})| + W(\mathbf{p})}, \\
W(\mathbf{p})~~&=~~max(1, W(\mathbf{p}) + W_t(\mathbf{p})).
\end{align}
When $|F_t(\mathbf{p}) - F(\mathbf{p})|$ exceeds the noise range $\epsilon_F$, we infer that the voxel region has changed. In this case, $W(\mathbf{p})$ decreases, and the reliability of the TSDF value is reduced. The map is updated for all voxels within the viewing frustum at each frame, allowing it to promptly adjust the TSDF values in response to semi-static changes in the scene. 

\subsubsection{3DGS}\label{sec:3dgs}
3DGS represents a scene using 3D Gaussian primitives, each parameterized by optimizable variables including the mean $\boldsymbol{\mu}$, covariance matrix $\mathbf{\Sigma}$ decomposed into scale matrix $\mathbf{S}$ and rotation $\mathbf{R}$, opacity $\alpha$, and view-dependent color $\mathbf{c}$ modeled with third-order spherical harmonics. To facilitate efficient pruning (as described in Sec.~\ref{sec:gadc}), each primitive additionally stores the Morton code $V$ of the voxel to which it belonged at initialization.

Given the camera pose $\mathbf{T}_t \mapsto [\mathbf{R}_t, \mathbf{t}_t]$ at frame $t$, each 3D Gaussian can be projected onto the image plane via splatting, resulting in a 2D Gaussian $\mathcal{N}(\boldsymbol{\mu_{\mathrm{I}}}, \mathbf{\Sigma_I})$ as follows:
\begin{equation}
\boldsymbol{\mu_{\mathrm{I}}}=\pi(\mathbf{T}^{-1}\cdot\boldsymbol{\mu}),~~
\mathbf{\Sigma_I} = \mathbf{JR}_t^{\mathrm{T}}\mathbf{(RS}\mathbf{S}^{\mathrm{T}}\mathbf{R}^{\mathrm{T}})\mathbf{R}_t\mathbf{J}^{\mathrm{T}},
\end{equation}
where $\mathbf{J}$ is the Jacobian of the affine approximation of the projective transformation that converts camera coordinates to ray coordinates. Based on the depth values $z$ of the 3D Gaussians in the camera coordinate system, the corresponding 2D Gaussians are blended, and the pixel color $\bar{\mathbf{I}}[\mathbf{u}]$ and depth $\bar{\mathbf{D}}[\mathbf{u}]$ are rendered as 
\begin{align}
\bar{\mathbf{I}}[\mathbf{u}] &= \sum_{i}^{N}\mathbf{c}_i\alpha_i\mathcal{N}_i(\mathbf{u}~|~\boldsymbol{\mu_{\mathrm{I}}}, \mathbf{\Sigma_I}) \prod_{j=1}^{i-1}(1 - \alpha_j\mathcal{N}_j(\mathbf{u}~|~\boldsymbol{\mu_{\mathrm{I}}}, \mathbf{\Sigma_I})),\\
\bar{\mathbf{D}}[\mathbf{u}] &= \sum_{i}^{N}z_i\alpha_i\mathcal{N}_i(\mathbf{u}~|~\boldsymbol{\mu_{\mathrm{I}}}, \mathbf{\Sigma_I}) \prod_{j=1}^{i-1}(1 - \alpha_j\mathcal{N}_j(\mathbf{u}~|~\boldsymbol{\mu_{\mathrm{I}}}, \mathbf{\Sigma_I})).
\end{align}

We optimize the 3D Gaussian primitives by backpropagating a loss constructed from rendered and captured images as follows:
\begin{equation}
\mathcal{L}_{rgb} = |\mathbf{I} - \bar{\mathbf{I}}|_1,~\mathcal{L}_{d} = |\mathbf{D} - \bar{\mathbf{D}}|_1.
\end{equation}
Unlike methods that only optimize the changed regions, our approach utilizes all image information for optimization. This helps avoid issues such as global illumination variations caused by the changed areas and missed regions resulting from inaccurate change detection. An up-to-date 3DGS map enables robots to obtain accurate photorealistic information in semi-static environments, which is crucial for high-level tasks such as image-goal navigation.

\subsection{Variation-aware Density Control}\label{sec:gadc}
In addition to the gradient-based optimization mentioned in Sec.~\ref{sec:3dgs}, another crucial step in optimizing 3D Gaussian primitives is adaptive density control (ADC). Most online mapping systems based on vanilla 3DGS first initialize Gaussian primitives by back-projecting downsampled depth maps, then increase or decrease their density through gradient-based densification or pruning. However, this adaptive mechanism has two limitations in semi-static scenes:
1) Without distinguishing between changed and unchanged regions, the initialization of 3DGS leads to redundant primitives in unchanged areas and insufficient primitives in changed areas, which significantly increases the burden of subsequent ADC and optimization processes.
2) gradient-based densification and pruning which require iterative backpropagation are slow to respond to significant map changes in real time. These limitations result in suboptimal performance and low efficiency in semi-static scenes. To address these issues, we propose variation-aware density control (VDC), which 1) replaces densification with an improved initialization scheme, and 2) directly prunes Gaussian primitives using geometric information from the hybrid map.

\subsubsection{Initialization}
Inspired by GSFusion~\cite{wei2024gsfusion}, we first segment the image using a quadtree based on the mean squared error of pixel values. The leaf nodes of the quadtree, corresponding to square image patches, are densely distributed in high-texture areas, while in low-texture regions, such as walls, the patches are relatively sparser and larger in size. Then, we traverse each leaf node to perform appearance-based and geometry-based variation detection.

For appearance-based variation detection (AVD), using the 3DGS prior map we render the image at the current camera pose and compute the local structural similarity index (SSIM) with captured image to measure the difference in appearance for each pixel at $\mathbf{u}$:
\begin{align}
&\mathrm{SSIM}(\mathbf{u}) = 
\frac{
(2\mu_{\bar{\mathbf{I}}} \mu_{\mathbf{I}} + C_1)
(2\Sigma_{\bar{\mathbf{I}}, \mathbf{I}} + C_2)
}{
(\mu_{\bar{\mathbf{I}}}^2 + \mu_\mathbf{I}^2 + C_1)
(\sigma_{\bar{\mathbf{I}}}^2 + \sigma_\mathbf{I}^2 + C_2)
}, \\
&\mu_{\mathbf{I}} = \frac{1}{|\mathcal{W}(\mathbf{u})|} 
\sum_{\mathbf{v} \in \mathcal{W}(\mathbf{u})} 
\mathbf{I}[\mathbf{v}], \\
&\sigma_{\mathbf{I}} = \left( 
\frac{1}{|\mathcal{W}(\mathbf{u})|-1}
\sum_{\mathbf{v} \in \mathcal{W}(\mathbf{u})}
(\mathbf{I}[\mathbf{v}] - \mu_\mathbf{I})^2 
\right)^{\frac{1}{2}}, \\
&\Sigma_{\bar{\mathbf{I}}, \mathbf{I}} = 
\frac{1}{|\mathcal{W}(\mathbf{u})|-1}
\sum_{\mathbf{v} \in \mathcal{W}(\mathbf{u})}
(\bar{\mathbf{I}}[\mathbf{v}] - \mu_{\bar{\mathbf{I}}})
(\mathbf{I}[\mathbf{v}] - \mu_{\mathbf{I}}),
\end{align}
where $\mathcal{W}(\mathbf{u})$ denotes a square patch of size $5 \times 5$ centered at $\mathbf{u}$. $C_1$ and $C_2$ are constants to avoid zero denominators.
We then calculate the mean SSIM value for each image patch. If the mean is below a threshold $\tau_s$ a sufficient inconsistency between the previous map and the current sensor data is detected, an indication of semi-static change. Notably, AVD not only detects areas where the appearance has changed but also identifies regions that are difficult for the current 3DGS map to reconstruct, leading to rendering quality improvements even in static scenes.

For geometry-based variation detection (GVD), we backproject the depth value of the center pixel of each image patch to obtain the corresponding 3D point. We then query the TSDF map using this 3D coordinate. If the voxel at that location has a weight greater than 1, no change in the region has happened and that voxel has already been initialized with a Gaussian primitive. In this case, no semi-static change has occurred. GVD prevents redundant initialization in the same space and focuses on regions affected by scene changes.

If either of the two detection passes, we initialize a Gaussian primitive for the image patch. 
The mean is initialized using the surface position $\mathbf{p}_c$ obtained by backprojecting the center coordinates $\mathbf{u}_c$ of the image patch. The low-frequency components of the spherical harmonics function are initialized using the RGB value at $\mathbf{u}_c$, with the remaining components set to zero. The opacity is set to 0.5 and rotation is identity matrix. For the initialization of scale matrix $\mathbf{S}$, we consider both image and geometric information: 
\begin{align}
\mathbf{S}&=d \cdot \mathrm{diag}(\frac{\mathbf{n}}{\|\mathbf{n}\|_2}),~~~d=\frac{L\cdot\mathbf{D}[\mathbf{u}_c]}{f_x},\\
\mathbf{n}&=\left\{
\begin{aligned}
&\mathbf{1}_{3} \oslash (\mathbf{1}_{3} + \mathrm{abs}(\nabla\mathbf{S}(\mathbf{p}_c))), &\nabla\mathbf{S}(\mathbf{p}_c) \neq \emptyset
,\\
&\mathbf{1}_{3}, &\mathrm{otherwise}.
\end{aligned}
\right.
\end{align}
where $f_x$ refers to the camera focal length, $L$ represents half of the side length of the image patch and $\oslash$ denotes element-wise division. $\nabla\mathbf{S}(\mathbf{p_c})$ denotes the surface normal at $\mathbf{p_c}$ by taking central finite-difference approximation of the TSDF values over neighboring voxels. $d$ is obtained by backprojecting $L$, reflecting the influence of image texture on the scale. 
The introduction of $\mathbf{n}$ biases the initialization of 3D Gaussians toward attaching flatly to the surface which reduces the number of optimization steps needed to achieve accurate alignment.
\subsubsection{Pruning}
Leveraging the ordered voxel structure of the hybrid map, we efficiently perform ray casting operations in the TSDF map to identify the voxels allocated between the camera's minimum measurable range $n_p$ and the measured depth $\mathbf{D}[\mathbf{u}]$ offest by a voxel size $s$:
\begin{align}
\mathbf{r}_{\mathbf{u}} = \Big\{ \mathbf{x}_t = z_t \big(\mathbf{K}^{-1}(\mathbf{u}^T|1)\big) ~\Big|~
   \begin{aligned}
      & z_t = n_p + t\cdot s, \\
      & t \in \mathbb{Z}_{\ge 0}, \\
      & z_t < \mathbf{D}[\mathbf{u}] - s
   \end{aligned}
\Big\}.
\end{align}
A TSDF value of a voxel in $\mathbf{r}_{\mathbf{u}}$ below a threshold $\tau_p$ indicates the presence of a deleted object in that region. By computing the Morton code of these voxels and matching them with the Morton codes stored during the initialization of the Gaussian primitives, we can efficiently identify the corresponding deleted areas and perform pruning on these Gaussians. The use of Morton code retrieval avoids the need for point-in-box checks, improving efficiency.

In addition, considering that localization errors and depth noise may cause voxels and Gaussian primitives to be allocated in regions not belonging to the actual surface during initialization, we identify voxels with TSDF values greater than 0.95 and prune their associated Gaussian primitives. Thanks to the noise-robust property of TSDFs, the values of these voxels quickly converge toward 1. By removing the corresponding Gaussian primitives in such regions, we effectively reduce floaters caused by noise. Although depth noise may occasionally lead to over-pruning, which results in degraded image quality in the affected regions, this issue is mitigated by the AVD module. Specifically, AVD detects these appearance degradations and re-initializes Gaussian primitives in the corresponding regions to restore rendering quality.

Unlike traditional 3DGS pruning, which occurs after optimization, our pruning operation takes place prior to the optimization of the current frame. Pruning outdated Gaussians based on geometric information significantly reduces the number of optimization steps and eliminates suboptimal results caused by outdated information, thereby ensuring accurate and timely map updates that are crucial for online robotic tasks such as localization.

\subsubsection{Keyframe Strategy}\label{sec:keyframe}
For each input frame, we first perform pruning and initialization. If the number of Gaussians that are deleted or added in the current frame exceeds a threshold $\tau_k$, the frame is marked as a keyframe, indicating that the newly observed information or scene change is substantial. To ensure that keyframes are evenly distributed along the trajectory, the current frame is also enforced as a keyframe if it is at least 10 frames apart from the last one. When a frame is designated as a keyframe, it is jointly optimized with a subset of randomly selected keyframes from the keyframe pool; otherwise, only the current frame contributes to the optimization of 3DGS.

%% file: chapter/experiments.tex
\section{Experiments}
\begin{figure*}[t]
    \centering
    \includegraphics[width=0.95\textwidth]{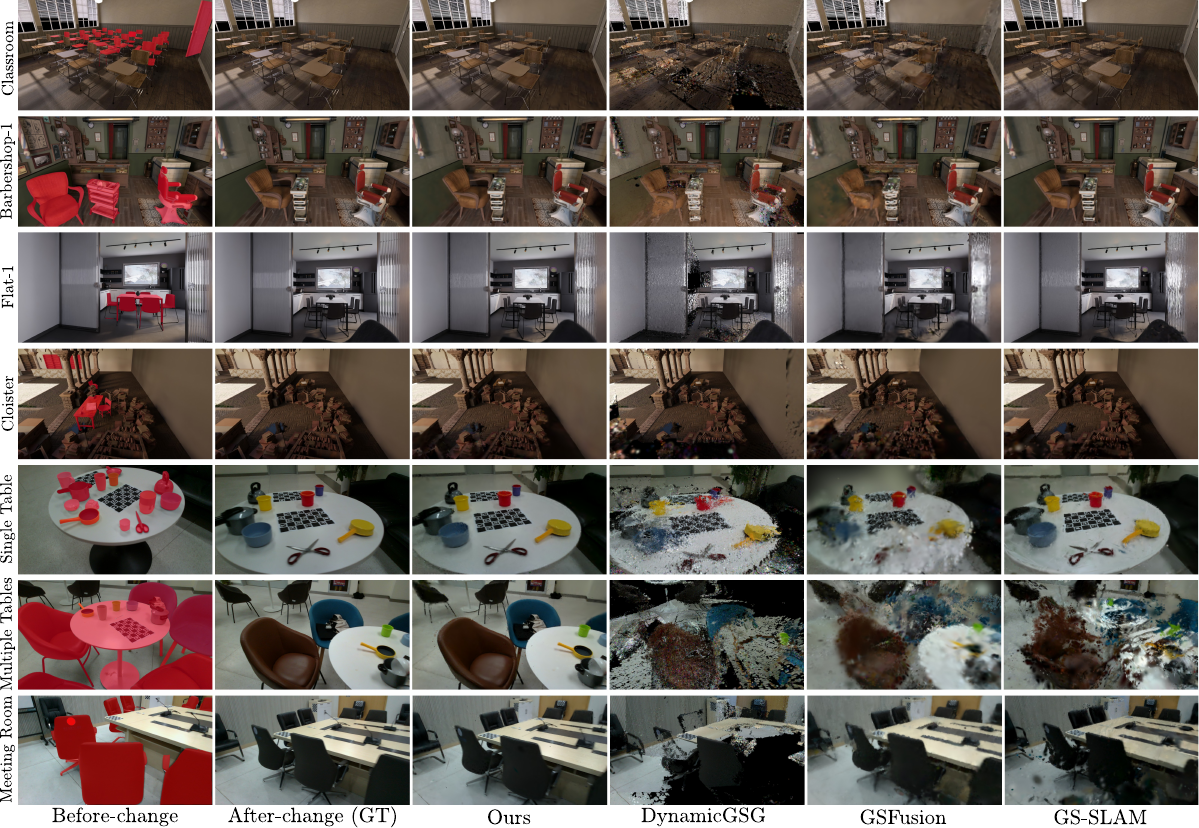}
    \caption{\textbf{Qualitative results on the VG-Scene dataset.}
    The first column shows the scene before changes, where the red masks highlight the objects or regions undergoing semi-static changes in the current view. The second column presents the observations after the changes. 
    Compared to the baselines, our method achieves higher-quality updates in the changed regions.
    }
    \label{fig:exp_show}
\end{figure*}
\subsection{Experiment Setup}
\subsubsection{Datasets}
Existing datasets~\cite{ackermann2025cl} used to evaluate 3DGS in semi-static scenes are designed for offline mapping systems. Due to their sparse, discontinuous viewpoints and lack of depth images, they are not suitable for assessing online RGB-D 3DGS mapping system. To resolve these issues and enable a comprehensive evaluation of our method, we construct an open-source dataset, VG-Scene, which includes both synthetic and real-world benchmarks. Each sequence consists of a pre-change and a post-change sub-sequence of the same scene. The synthetic benchmark is built on publicly available Blender demo scenes, where we define camera trajectories and introduce object changes, resulting in six sequences in total. The real-world benchmark consists of three sequences ranging from table-scale to room-scale, captured using an Intel RealSense L515.
The VG-Scene dataset covers common types of scene changes, including object addition, removal, and rearrangement. In addition, we conduct experiments on four ScanNet++ sequences (8b5caf3398, 39f36da05b, b20a261fdf, and f34d532901), following the same protocol as GS-Fusion, to further evaluate the applicability of our method in static environments.

\subsubsection{Baselines}
We compare our method with state-of-the-art RGB-D online mapping systems, namely GS-SLAM~\cite{matsuki2024gaussian}, GSFusion~\cite{wei2024gsfusion} and DynamicGSG~\cite{ge2025dynamicgsg}. 
For fair comparison, We disable the tracking modules in both GS-SLAM and DynamicGSG and use the robot location information from VINS-Mono~\cite{qin2018vins}. During evaluation, we record the camera trajectories estimated by VINS-Mono~\cite{qin2018vins} and reuse them across all methods.

\subsubsection{Metrics}
Following common practice, we report geometric accuracy using Abs Rel (relative $l_1$ error) and $\delta_1$ (percentage of correctly estimated pixels within a thresh of 1.25)~\cite{eigen2014depth}, and evaluate rendering quality using PSNR, SSIM, and LPIPS.
In addition, we measure the mapping thread’s runtime frame rate to assess computational efficiency. Best and second best results in the tables are boldfaced and underlined, respectively.
\subsubsection{Implementation Details}
All experiments were conducted on a single NVIDIA A6000 GPU due to the substantial memory requirements of GS-SLAM. Our system is implemented in C++ with LibTorch, while 3DGS rasterization and optimization, and quadtree are accelerated with CUDA. For all sequences, we employ a unified parameter configuration: the voxel size is set to 0.01 m, the threshold for appearance-based variation detection is $\tau_s = 0.6$, the pruning threshold is $\tau_p = 0.2$, and the keyframe selection threshold is $\tau_k = 200$.

\subsection{Map Update Performance}
\begin{table*}[t]
\caption{\textbf{Quantitative Comparison on VG-Scene dataset}. The metrics are averaged over all sequences.}
\renewcommand{\arraystretch}{1.3}
\begin{center}
\begin{tabular}{c cccccc cccccc}
\toprule
& \multicolumn{6}{c}{\textbf{Synthetic}} & \multicolumn{6}{c}{\textbf{Real World}} \\
\cmidrule(lr){2-7} \cmidrule(lr){8-13}
\textbf{Method} & \textbf{PSNR}$\uparrow$ & \textbf{SSIM}$\uparrow$ & \textbf{LPIPS}$\downarrow$ & \textbf{Abs Rel}$\downarrow$ & $\boldsymbol{\delta_1}\uparrow$ & \textbf{FPS}$\uparrow$
       & \textbf{PSNR}$\uparrow$ & \textbf{SSIM}$\uparrow$ & \textbf{LPIPS}$\downarrow$ & \textbf{Abs Rel}$\downarrow$ & $\boldsymbol{\delta_1}\uparrow$ & \textbf{FPS}$\uparrow$ \\
\midrule
GS-SLAM~\cite{matsuki2024gaussian} & 25.94 & 0.84 & 0.12 & 0.026 & 0.97 & 2.31 & 18.52 & 0.68 & 0.37 & 0.096 & 0.86 & 2.7 \\
GSFusion~\cite{wei2024gsfusion} & 28.41 & 0.88 & 0.15 & 0.089 & 0.87 & \textbf{9.30} & 21.75 & 0.77 & 0.37 & 0.150 & 0.74 & \textbf{7.76} \\
DynamicGSG~\cite{ge2025dynamicgsg} & 22.08 & 0.71 & 0.29 & 0.054 & 0.93 & 0.18 & 13.54 & 0.38 & 0.57 & 0.246 & 0.72 & 0.13 \\
Ours     & \textbf{33.14} & \textbf{0.94} & \textbf{0.06} & \textbf{0.013} & \textbf{0.99} & \underline{7.47} & \textbf{24.21} & \textbf{0.84} & \textbf{0.26} & \textbf{0.072} & \textbf{0.90} & \underline{5.86} \\
\bottomrule
\end{tabular}
\end{center}
\label{tab:gs3data}
\vspace*{-0.18in}
\end{table*}

\subsubsection{Semi-static Scene}
We evaluate our method on the VG-Scene dataset to assess its performance in semi-static scenes. All methods follow the same experimental protocol in which the pre-change subsequence is used to construct the prior map, and the post-change subsequence is then used to update it. Quantitative results are summarized in Tab.~\ref{tab:gs3data}, where our approach significantly outperforms baselines in rendering quality. 
Compared with DynamicGSG, our method improves PSNR by approximately 50\% while achieving about a 40$\times$ increase in frame rate.
Furthermore, owing to accurate map updates, the rendered depth exhibits the lowest error.
Fig.~\ref{fig:exp_show} presents qualitative comparisons, showing that our decoupling method produces high-fidelity updates in changed regions, even under substantial scene changes or in real-world scenarios with noisy depth measurements.
Experiments show that updating changed regions through optimization-based schemes makes the optimization process harder to converge, while object-level local update strategies are prone to over-deletion or incomplete updates.

\begin{table}[t]
\caption{\textbf{The Comparisons of Rendering results on ScanNet++}.}
\renewcommand{\arraystretch}{1.3}
\begin{center}
\begin{tabular}{cccccc}
\hline
                               & \textbf{Method}   & \textbf{PSNR}$\uparrow$ & \textbf{SSIM}$\uparrow$ & \textbf{LPIPS}$\downarrow$ & \textbf{FPS}  \\ \hline
\multirow{2}{*}{Training view} & GSFusion & 27.97          & 0.87           & 0.15              & \textbf{9.17} \\
                               & Ours     & \textbf{28.42} & \textbf{0.88}  & \textbf{0.10}     & 8.87 \\ \hline
\multirow{2}{*}{Novel view}    & GSFusion & 24.96          & 0.83           & 0.23              & $\times$    \\
                               & Ours     & \textbf{25.36} & \textbf{0.84}  & \textbf{0.17}     & $\times$    \\ \hline
\end{tabular}
\end{center}
\label{tab:scannetpp}
\end{table}

\subsubsection{Static Scene}
We further evaluate the rendering quality of our method on static scenes using the ScanNet++ dataset, with results reported in Tab.~\ref{tab:scannetpp}. Although our framework is primarily designed for semi-static environments, it also brings improvements in static settings. This is attributed to the appearance-based variation detection, which not only identifies dynamic changes but also captures regions that are poorly fitted by the current map. In addition, the pruning stage removes Gaussians that are far from the surface, effectively reducing floating artifacts. Together, these mechanisms contribute to enhanced rendering quality even in fully static scenes.

\begin{table}[t]
\renewcommand{\arraystretch}{1.3}
\caption{\textbf{Ablation Studies on VG-Scene dataset}. We test the impact of pruning, AVD, GVD, SSIM threshold $\tau_s$ and pruning threshold $\tau_p$.}
\begin{center}
\begin{tabular}{ccc|cccc}
\hline
            & $\tau_s$ & $\tau_p$ & \textbf{PSNR}$\uparrow$ & \textbf{SSIM}$\uparrow$ & \textbf{LPIPS}$\downarrow$ & \textbf{FPS}$\uparrow$ \\ \hline
w/o pruning & 0.6      & N/A     & 26.90          & 0.857          & 0.21              & 6.88          \\
w/o AVD     & N/A      & 0.4      & 25.59          & 0.842          & 0.24              & \textbf{8.30}          \\
w/o GVD     & 0.6      & 0.4      & 27.44          & 0.864          & 0.20              & 7.61          \\ \hline
Ours        & 0.4      & 0.4      & 25.92          & 0.849          & 0.242             & \underline{7.80}          \\
Ours        & 0.7      & 0.4      & \textbf{29.23} & \textbf{0.885} & \textbf{0.163}    & 5.35          \\ \hline
Ours        & 0.6      & 0.2      & 28.48          & 0.873          & 0.187             & 6.63          \\
Ours        & 0.6      & 0.6      & 28.62          & 0.873          & 0.187             & 6.49          \\ \hline
Ours        & 0.6      & 0.4      & \underline{28.66} & \underline{0.877} & \underline{0.187}    & 6.67 \\ \hline
\end{tabular}
\end{center}
\label{tab:abla}
\vspace*{-0.18in}
\end{table}

\begin{figure}[t]
    \centering
    \includegraphics[width=(0.95\textwidth-0.95\columnsep)/2]{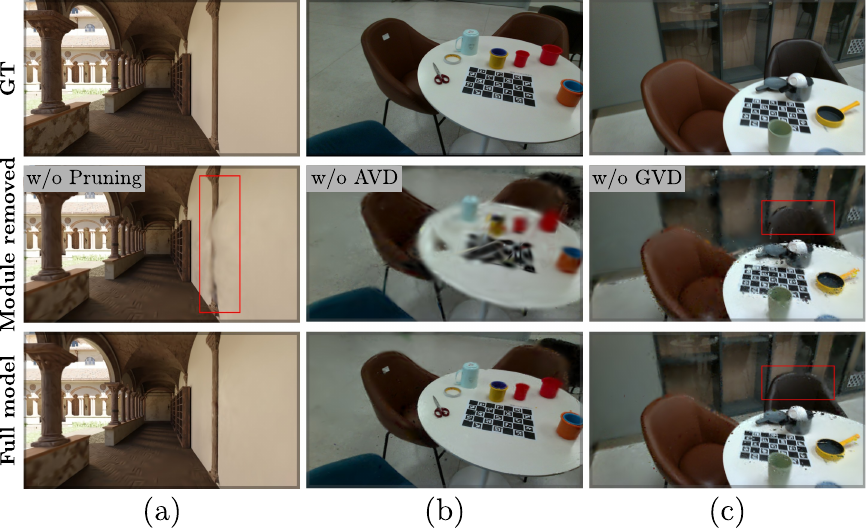}
    \caption{\textbf{Effectiveness analysis of each component}.
    (a) Pruning reduces the influence of outdated Gaussians during optimization, preventing suboptimal results. (b) AVD facilitates the detection of appearance changes without geometric displacement, while (c) GVD resolves ambiguities in regions with similar appearance but different depth.
    }
    \label{fig:ablation}
    \vspace*{-0.18in}
\end{figure}

\subsection{Ablation Studies}
We conduct ablation studies on the initialization, pruning modules using three sequences ($Classroom$, $Cloister$, and $Single~table$) from the VG-Scene dataset. The results are summarized in Tab~\ref{tab:abla}.
\subsubsection{Effect of Pruning}
Our pruning strategy efficiently removes outdated Gaussians prior to optimization, leading to improved rendering quality. As illustrated in Fig.~\ref{fig:ablation} (a), the benefits are particularly notable in changed regions with limited observations. Moreover, by directly storing Morton codes and avoiding expensive point-in-box checks, our pruning method incurs minimal computational overhead.

\subsubsection{Effect of Initialization}
The results in the second, third, and last rows of Tab.~\ref{tab:abla} demonstrate that both AVD and GVD significantly enhance the quality of map updates. AVD facilitates the detection of changes that do not involve geometric displacement, while GVD resolves ambiguities in regions with similar appearance but different depth, as illustrated in Fig.~\ref{fig:ablation} (b) and (c). The focused initialization in changed regions increases the number of Gaussian primitives to some extent, which in turn leads to a reduction in system frame rate.

\subsubsection{Effect of SSIM Threshold}
When the SSIM threshold $\tau_s$ increases, more pixels pass the AVD detection per frame, resulting in the initialization of a larger number of Gaussians. This enhances rendering quality, but the increased number of Gaussians in the scene also extends the optimization time, thereby lowering the system frame rate.

\subsubsection{Effect of Pruning Threshold}
As shown in the last three rows of Tab.~\ref{tab:abla}, varying the threshold $\tau_p$ from 0.2 to 0.6 has little impact on rendering quality, indicating that the pruning module is insensitive to $\tau_p$. While a higher threshold results in more voxels being marked as changed, our retrieval strategy keeps the computational overhead nearly unchanged.

\subsection{Applications}
Our research is motivated by the desire to provide an accurate and photorealistic map capable of handling semi-static changes to support downstream robot tasks. In this section, we evaluate the effectiveness of our constructed maps in two representative downstream tasks: open-vocabulary segmentation and 6D pose estimation.

\subsubsection{Open-vocabulary Segmentation}
\begin{figure}[t]
    \centering
    \includegraphics[width=(0.95\textwidth-0.95\columnsep)/2]{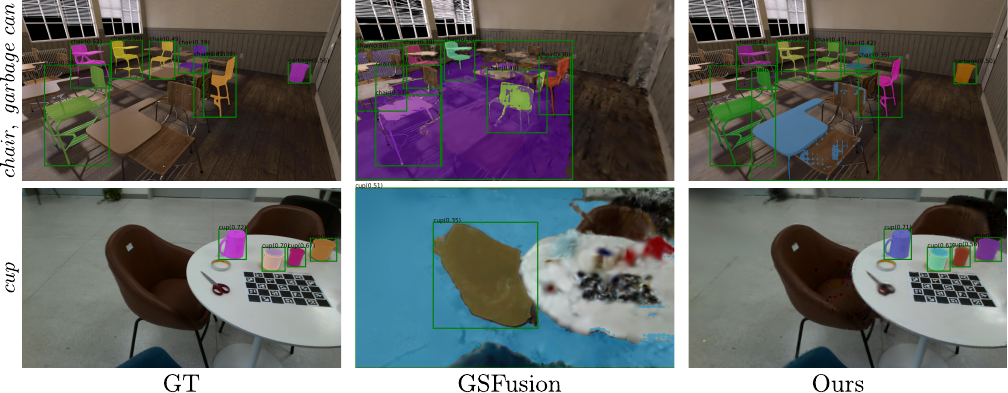}
    \caption{\textbf{Qualitative results of Grounded-SAM on rendered images}. The left-hand text denotes the prompt. We visualize the masks, bounding boxes, and confidence produced by Grounded-SAM.}
    \label{fig:sam}
\end{figure}

Owing to the capability of 3DGS to render photorealistic images from arbitrary viewpoints, several studies have leveraged 3DGS as a bridge between 3D maps and 2D visual foundation models (e.g., DINO~\cite{oquab2023dinov2}, SAM~\cite{kirillov2023segment}) to support tasks such as image-goal navigation and manipulation. A critical precondition is that the rendered images serve as effective inputs to visual foundation models. We employ Grounded-SAM~\cite{ren2024grounded} to perform open-vocabulary segmentation on the rendered images and conduct a qualitative evaluation, as illustrated in Fig.~\ref{fig:sam}. Benefiting from the precise scene-change updates achieved by our method in semi-static environments, the rendered images produced by our method lead to much better segmented objects than the competing method of GSFusion.

\subsubsection{6D Pose Estimation}
Recent studies~\cite{matteo20246dgs,cheng2025logs} have leveraged the differentiability and novel-view rendering capabilities of 3DGS maps to perform global pose estimation.  We use the 3DGS maps reconstructed by GSFusion and by our method as inputs to 6DGS~\cite{matteo20246dgs} for global pose estimation, and compare them using mean angular (MAE) and mean translation (MTE) error. Experiments are conducted on the VG-Scene dataset using two synthetic sequences and one real-world sequence. For each synthetic sequence, ten test images are rendered outside the training trajectory using Blender, while for the real sequence, ten test images are uniformly sampled from the post-change sequence. As shown in Tab.~\ref{tab:6dgs}, our method achieves lower pose estimation errors compared with the baseline, demonstrating its ability to accurately update maps in semi-static scenes and thereby better support downstream robotic tasks.

\begin{table}[t]
\renewcommand{\arraystretch}{1.3}
\caption{\textbf{Comparison of Pose Estimation}.}
\begin{center}
\begin{tabular}{c|cc|cc|cc}
\hline
 & \multicolumn{2}{c}{Classroom} & \multicolumn{2}{c}{Flat-2} & \multicolumn{2}{c}{Multiple Tables} \\ \cmidrule(lr){2-3} \cmidrule(lr){4-5} \cmidrule(lr){6-7}
                      \textbf{Method}  & \textbf{MAE}            & \textbf{MTE}           & \textbf{MAE}           & \textbf{MTE}         & \textbf{MAE}             & \textbf{MTE}            \\ \cmidrule(){0-0} \cmidrule(lr){2-3} \cmidrule(lr){4-5} \cmidrule(lr){6-7}
GSFusion                & 58.20          & 1.57          & 54.94         & 1.10        & 37.19           & 0.28           \\
Ours                    & \textbf{24.02}          & \textbf{0.79}          & \textbf{34.48}         & \textbf{0.85}        & \textbf{29.03}           & \textbf{0.18}           \\ \hline
\end{tabular}
\end{center}
\label{tab:6dgs}
\vspace*{-0.18in}
\end{table}

%% file: chapter/conclusion.tex
\section{Conclusions}
We propose an online RGB-D 3DGS mapping system tailored for semi-static environments. Our approach builds a hybrid map representation by combining TSDF and 3DGS. By exploiting the geometric information of the TSDF map, we introduce a variation-aware density control module that enables the system to perform proper update of regions experiencing semi-static changes {as well as inadequately reconstructed areas}. Furthermore, we construct a semi-static scene dataset to evaluate RGB-D 3DGS online mapping. Experimental results demonstrate that our system achieves high-quality and efficient map updates in semi-static environments, and provides up-to-date 3DGS maps that are highly beneficial for downstream robotic tasks.

In future work, we plan to extend the proposed mapping system into a full SLAM framework and support multiple types of depth sensors. Moreover, we will extend VG-Scene to larger-scale outdoor environments, enabling the study of semi-static changes in more complex and realistic scenarios.